\documentclass[conference]{IEEEtran}
\IEEEoverridecommandlockouts
\usepackage{cite}
\usepackage{amsmath,amssymb,amsfonts}
\usepackage{algorithmic}
\usepackage{graphicx}
\usepackage{textcomp}
\usepackage{xcolor}
\usepackage{booktabs}
\usepackage{multirow}
\usepackage{array}
\usepackage{url}
\usepackage{hyperref}
\usepackage{float}
\usepackage{balance}
\usepackage{eso-pic}
\usepackage{microtype} 

\AddToShipoutPictureFG{%
  \ifnum\value{page}=1
    \AtPageUpperLeft{%
      \hspace{0.68in}%
      \raisebox{-0.40in}{\parbox{7in}{\fontsize{9}{9.2}\selectfont
        2026 1st International Conference on Next-Generation Electrical \& Electronics, Computer Systems, and Technologies (iCONEECT)\\
        25--26 September, 2026, Premier University, Chittagong, Bangladesh}}%
    }%
    \AtPageLowerLeft{%
      \hspace{0.68in}%
      \raisebox{0.30in}{\fontsize{8}{9.2}\selectfont 979-8-3195-4810-8/26/\$31.00~\copyright 2026 IEEE}%
    }%
  \fi
}

\hypersetup{
    colorlinks=false,
    hidelinks
}

\def\BibTeX{{\rm B\kern-.05em{\sc i\kern-.025em b}\kern-.08em
    T\kern-.1667em\lower.7ex\hbox{E}\kern-.125emX}}
\begin{document}

\title{Transfer Learning with Conformalized Quantile Regression for Solar PV Forecasting Under Load-Shedding-Driven Data Scarcity\\
\thanks{}
}

\author{\IEEEauthorblockN{1\textsuperscript{st} Rakib Abdullah}
\IEEEauthorblockA{\textit{Department of AI and Data Science} \\
\textit{Green University of Bangladesh}\\
rakib@ads.green.edu.bd}
\and
\IEEEauthorblockN{2\textsuperscript{nd} K.M.Tahlil Mahfuz Faruk}
\IEEEauthorblockA{\textit{Department of Software Engineering} \\
\textit{Green University of Bangladesh}\\
tahlil@swe.green.edu.bd}
}

\maketitle

\begin{abstract}

Solar photovoltaic (PV) power forecasting is central to the economic dispatch of battery-backed microgrids, yet two compounding challenges make it particularly hard in many developing-world deployments: extreme data scarcity at newly commissioned sites, and structured measurement gaps introduced by load-shedding-driven grid outages. We address both challenges for a representative 10~kWp rooftop microgrid in Bangladesh by combining transfer learning with conformal uncertainty quantification. A long short-term memory (LSTM) network is first pretrained on 148{,}537 hourly observations from the DKASC Alice Springs photovoltaic dataset and then fine-tuned on one and three months of pvlib-simulated Bangladesh target data to which a Bangladesh Power Development Board (BPDB)-derived outage mask has been applied before training. Transfer learning reduces point-forecast RMSE by 23.7\% under one-month scarcity (0.503 vs.\ 0.659~kW) and by 13.7\% under three-month scarcity (0.421 vs.\ 0.488~kW). Critically, the one-month scratch-trained model achieves a skill score of $-$0.033 relative to a 24~h persistence baseline, meaning it is marginally worse than simply repeating yesterday's output, while the transfer counterpart reaches $+$0.212. Conformalized quantile regression (CQR) is then applied post-hoc to each model. At three months of target data, the Scratch+CQR model undercovers at 84.2\%, violating the 90\% nominal guarantee, whereas Transfer+CQR achieves 94.3\% empirical coverage with 14\% narrower prediction intervals (1.98 vs.\ 2.31~kW mean width). We argue that pretrained source-domain representations stabilise both the point predictor and the conformalization calibration set under outage-driven missingness, an effect that diminishes gracefully as target data grow.

\end{abstract}

\section{Introduction}

The global energy transition has placed solar photovoltaics at the centre of plans to extend 
reliable electricity access to the roughly 770 million people who currently lack it 
\cite{iea2023world,irena2023renewable,un2015sdg}. Battery-backed microgrids represent perhaps 
the most pragmatic near-term pathway to SDG~7 in regions with weak or unreliable national grids 
\cite{hirsch2018microgrids,ton2012department,guerrero2011hierarchical}. Bangladesh is an instructive case: the country 
has a rapidly growing rooftop solar sector, yet its distribution network still experiences 
pronounced load-shedding, particularly during the summer peak-demand season, with outage 
frequencies that routinely exceed 30\% of daylight hours in some districts 
\cite{hossain2020nexus,mondal2011renewable}. For a microgrid energy-management system (EMS), 
accurate next-day PV power forecasts directly translate into decisions about when to charge 
batteries and when to start or stop a diesel backup generator; errors in those forecasts carry 
real fuel costs and sometimes real welfare costs when loads are shed unnecessarily.

The forecasting problem is doubly hard here. First, a newly installed system has very 
little local historical data; transfer learning-adapting a model pretrained on a data-rich 
source to a data-scarce target-is the natural remedy 
\cite{pan2010survey,yosinski2014transferable,sarmas2022transfer}, but prior work largely 
assumes continuous, clean target training data. Second, load-shedding leaves structured 
gaps: measurement systems go offline precisely during high-demand hours, a non-random 
missingness pattern that can corrupt both model training and the calibration sets 
underlying conformal coverage guarantees \cite{romano2019conformalized,tibshirani2019conformal}.

This intersection has received little systematic attention. Probabilistic PV forecasting 
\cite{vandermeer2018review,hong2016probabilistic} and conformal prediction 
\cite{stankeviciute2021conformal,renkema2024conformal,moradi2025enhanced} are both well 
studied, and transfer learning for solar forecasting has seen a flurry of activity 
\cite{sarmas2022transfer,weiss2016survey}, but no work carefully examines how these methods 
interact under outage-driven structured data gaps, or whether combining them yields a 
reliable system under the severe data budgets realistic for a newly deployed Bangladesh 
microgrid.

This paper makes four concrete contributions:

\begin{enumerate}
    \item A reproducible data pipeline that synthesises a realistic Bangladesh PV dataset 
    from NASA POWER meteorological reanalysis via pvlib \cite{holmgren2018pvlib}, 
    then applies a BPDB-derived seasonal outage mask to introduce structured missingness 
    before training.

    \item An empirical comparison of four LSTM-based forecasting models-trained from 
    scratch or fine-tuned from a DKASC-pretrained backbone, under one-month and three-month 
    target data budgets-evaluated on a clean December 2023 hold-out set of 744 hourly 
    observations.

    \item A conformal quantile regression (CQR) layer applied post-hoc to each model, 
    with an analysis of how outage-driven covariate shift degrades conformal coverage 
    guarantees for scratch-trained models but not for transfer-initialised ones.

    \item An hourly coverage diagnostic that reveals which parts of the solar day are 
    most affected by structured missingness, providing actionable guidance for EMS design.
\end{enumerate}

The remainder of the paper is structured as follows. Section~\ref{sec:related} reviews 
relevant prior work. Section~\ref{sec:methodology} describes the data pipeline, model 
architecture, and calibration procedure. Section~\ref{sec:experiments} presents 
experimental results. Section~\ref{sec:discussion} interprets the findings, and 
Section~\ref{sec:conclusion} concludes.

\section{Related Work}
\label{sec:related}

Recurrent neural networks, particularly LSTMs \cite{hochreiter1997lstm}, are the default 
choice for short-term PV and irradiance forecasting 
\cite{diagne2013review,inman2013solar,blaga2019current,wang2019comparison,antonanzas2016review}, 
typically outperforming autoregressive and physical-model baselines by learning temporal 
dependencies across lagged meteorological and power features. Data scarcity has pushed 
researchers toward cross-site transfer learning \cite{sarmas2022transfer}, but most 
benchmarks assume clean, contiguous training records; we extend this line of work to the 
more realistic case of non-random gaps, which, as far as we can tell, has not been 
explicitly controlled for in the transfer-learning literature.

This question of transfer sits within a broader literature on transfer learning for 
energy systems. Pan and Yang's survey \cite{pan2010survey} remains the standard reference 
for the domain/task-adaptation distinction, and representation-learning advances 
\cite{bengio2013representation,lecun2015deep} show that features learned on one domain 
often generalise to related ones-the theoretical basis for the cross-site transfer studied 
here. Yosinski et al.\ \cite{yosinski2014transferable} examined how well features transfer 
between layers and where fine-tuning should begin; for PV specifically, Sarmas et al.\ 
\cite{sarmas2022transfer} showed pretrained networks can adapt to data-scarce sites with 
just weeks of local data, while Weiss et al.\ \cite{weiss2016survey} survey the risk of 
negative transfer when source and target domains are poorly matched. Our source (Alice 
Springs, 23.7°S, arid) and target (Dhaka, 23.8°N, humid subtropical monsoon) sit at 
opposite ends of that spectrum, making beneficial-versus-harmful transfer a real empirical 
question rather than a formality.

Beyond point forecasts, calibrated prediction intervals are now seen as necessary for 
robust energy-system optimisation \cite{gneiting2007strictly,hong2016probabilistic,pinson2013wind}, 
with quantile regression, Gaussian processes, and ensembles as classical tools 
\cite{vandermeer2018review}. Conformal prediction 
\cite{vovk2005algorithmic,shafer2008tutorial,fontana2023conformal} offers a 
distribution-free alternative, yielding finite-sample coverage guarantees under 
exchangeability regardless of the underlying model. Conformalized quantile regression 
\cite{romano2019conformalized} is especially efficient when the base quantile regressor is 
well calibrated, and Tibshirani et al.\ \cite{tibshirani2019conformal} extended the 
approach to covariate shift-relevant here, since outages distort the training distribution 
relative to the clean test set. Conformal prediction has also reached renewable 
forecasting: Renkema et al.\ \cite{renkema2024conformal} applied CQR to electricity-market 
PV forecasting, Moradi et al.\ \cite{moradi2025enhanced} conditioned the conformal score on 
meteorological regime for tighter clear-sky intervals, and Stankeviciute et al.\ 
\cite{stankeviciute2021conformal} extended exchangeability to multi-step, autoregressive 
forecasting. None of this work, however, asks what happens to CQR calibration when the 
calibration set itself carries load-shedding-driven structured missingness-the question we 
take up here.

\section{Proposed Methodology}
\label{sec:methodology}
\begin{figure}[h]
    \centering
    \includegraphics[width=0.85\columnwidth]{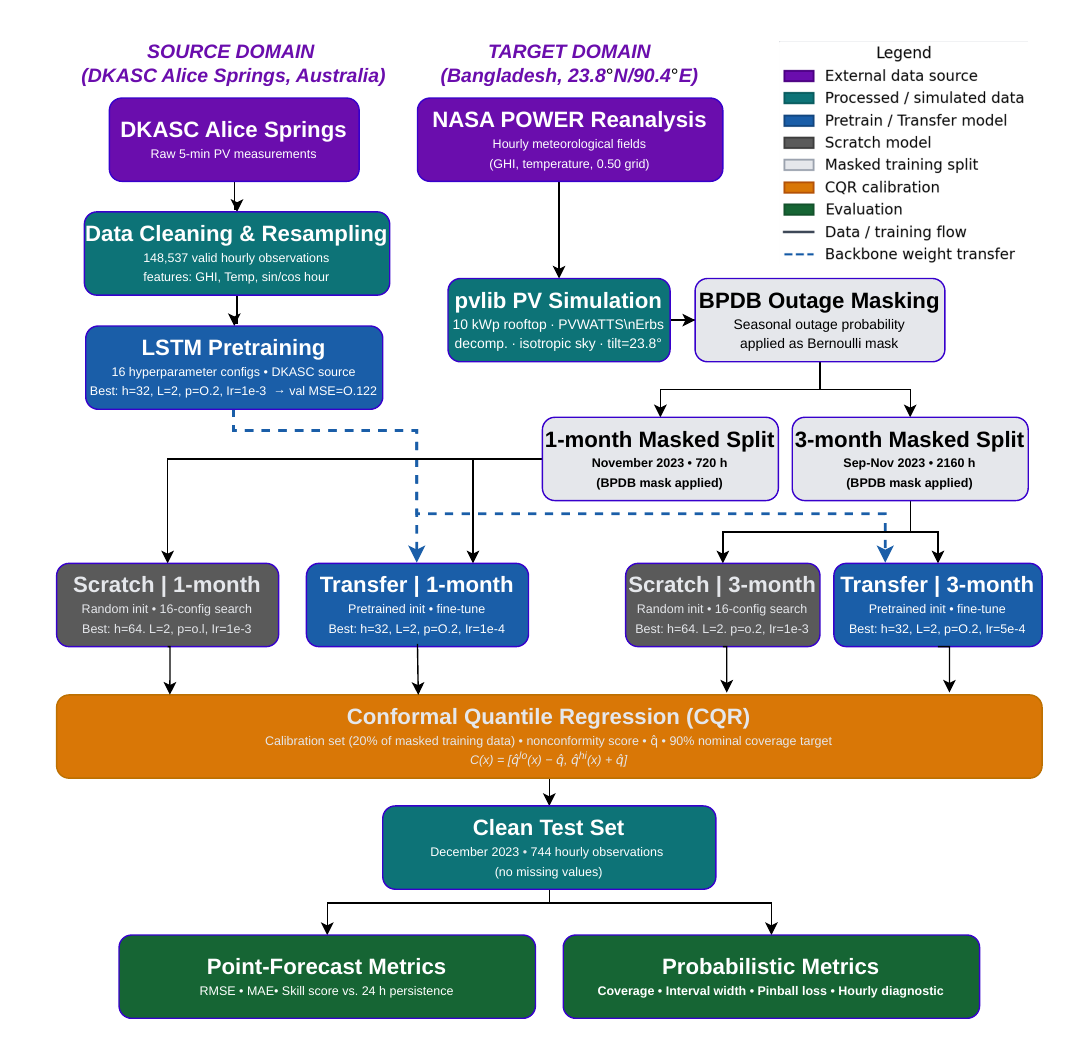}      
    \caption{Full methodology pipeline from DKASC source pretraining and BPDB-masked Bangladesh target splits through four LSTM variants, CQR calibration, and final evaluation on the December 2023 hold-out set.} 
    \label{fig:microgrid_pipeline}
\end{figure}

\subsection{Source-Domain Data: DKASC Alice Springs}

The source dataset comes from the Desert Knowledge Australia Solar Centre (DKASC) at 
Alice Springs (23.7°S, 133.9°E), a large-scale outdoor photovoltaic testing facility. 
After filtering for the C-Phase of Site 106 and resampling native 5-minute readings 
to hourly averages, we retain 148{,}537 valid observations spanning approximately 
seventeen years. Features used are global horizontal irradiance (GHI), ambient 
temperature, and cyclically encoded time-of-day (sine/cosine hour encoding); the 
target variable is AC power output in kilowatts. This dataset was chosen for pretraining 
for its length and data quality, and because rooftop silicon PV in a semi-arid, 
high-irradiance environment shares clear-sky ramp characteristics with tropical sites 
even when absolute irradiance and seasonal patterns differ. DKASC data are publicly 
available through the Australian Renewable Energy Agency (ARENA) open data portal.

\subsection{Target-Domain Data: Simulated Bangladesh PV}

The target site is a hypothetical 10~kWp rooftop crystalline silicon system at 
23.8°N, 90.4°E (Dhaka region). Hourly weather data for 2023 were retrieved from 
the NASA POWER reanalysis product \cite{sengupta2018national}, which provides 
globally gridded surface meteorological fields at 0.5° resolution. PV generation 
was then simulated using pvlib \cite{holmgren2018pvlib} with the PVWATTS performance model, the Erbs decomposition model to separate direct and diffuse irradiance components, and plane-of-array irradiance calculated via the isotropic sky model at a fixed, latitude-equal surface tilt. 
The resulting 8{,}760-hour dataset is a physically consistent surrogate for a real 
installed system, avoiding reliance on ground measurements that do not exist for such a 
site and allowing clean experimental control.

\subsection{Structured Missingness Model}

A distinguishing feature of this work is our explicit treatment of load-shedding-driven 
data gaps. An hour-of-day outage probability vector was derived from BPDB load dispatch 
statistics, which show that outage frequency peaks during afternoon demand hours 
(13:00--17:00 local time, $p \approx 0.33$) and is lowest at night. This probability 
vector is applied as a Bernoulli mask to the training splits before any model sees the 
data. The mask is fixed once, before any experiment, to avoid information leakage. 
The test set (December 2023) is kept entirely clean to reflect the assumption that 
a monitoring system logs data even during outages when running from battery backup.

\subsection{Training Splits and Evaluation Protocol}

Two target training splits are formed: a one-month split (November 2023, 720 observations) 
and a three-month split (September--November 2023, 2{,}184 observations). After applying 
the outage mask, 80\% of each masked split is used for gradient updates and the remaining 
20\% serves as the CQR calibration set. The hold-out test set (December 2023, 744 
observations) is always clean.

\subsection{LSTM Architecture and Training}

All models share a common LSTM backbone \cite{hochreiter1997lstm,lecun2015deep} followed by a dense output head. Although gated recurrent units (GRUs) \cite{cho2014learning} offer a lighter alternative, LSTMs were preferred here because their two-gate structure better retains long-range diurnal dependencies across the 24-hour input window. The architecture 
is parameterised by hidden size $h \in \{32, 64\}$, number of stacked layers $L \in \{1, 2\}$, 
and dropout probability $p_d \in \{0.1, 0.2\}$. Inputs are presented as sliding windows 
of length 24 (one solar day) with the four engineered features described above. The 
point-forecast head outputs a single value; a separate quantile head trained with the 
pinball loss at $\alpha \in \{0.05, 0.95\}$ provides the pre-conformal quantile bounds 
used by CQR.

\textbf{Source pretraining.} A grid search over 16 hyperparameter combinations was 
conducted on the DKASC dataset. The best configuration (hidden size 32, 2 layers, 
dropout 0.2, learning rate $10^{-3}$, optimised with Adam \cite{kingma2015adam}) achieved 
a validation MSE of 0.123~kW$^2$ on a held-out DKASC test year. This model's weights 
serve as the initialisation point for all Transfer experiments.
Table~\ref{tab:hparam_best} reports the selected best configuration for each model tag.


\begin{table}[htbp]
\caption{Best Hyperparameter Configuration per Model Tag}
\begin{center}
\begin{tabular}{|l|c|c|c|c|}
\hline
\textbf{Tag} & \textbf{$h^{\mathrm{a}}$} & \textbf{$L^{\mathrm{b}}$} & \textbf{$p_d^{\mathrm{c}}$} & \textbf{Val MSE} \\
\hline
Pretrain (DKASC) & 32 & 2 & 0.20 & 0.1226 \\
\hline
Scratch 1-month  & 64 & 2 & 0.10 & 0.6252 \\
\hline
Transfer 1-month & 32 & 2 & 0.20 & 0.5751 \\
\hline
Scratch 3-month  & 64 & 2 & 0.20 & 0.5090 \\
\hline
Transfer 3-month & 32 & 2 & 0.20 & 0.4504 \\
\hline
\multicolumn{5}{l}{$^{\mathrm{a}}h$: Hidden size, $^{\mathrm{b}}L$: Number of LSTM layers, $^{\mathrm{c}}p_d$: Dropout probability.} \\
\end{tabular}
\label{tab:hparam_best}
\end{center}
\end{table}

\begin{figure}[h]
    \centering
    \includegraphics[width=0.85\columnwidth]{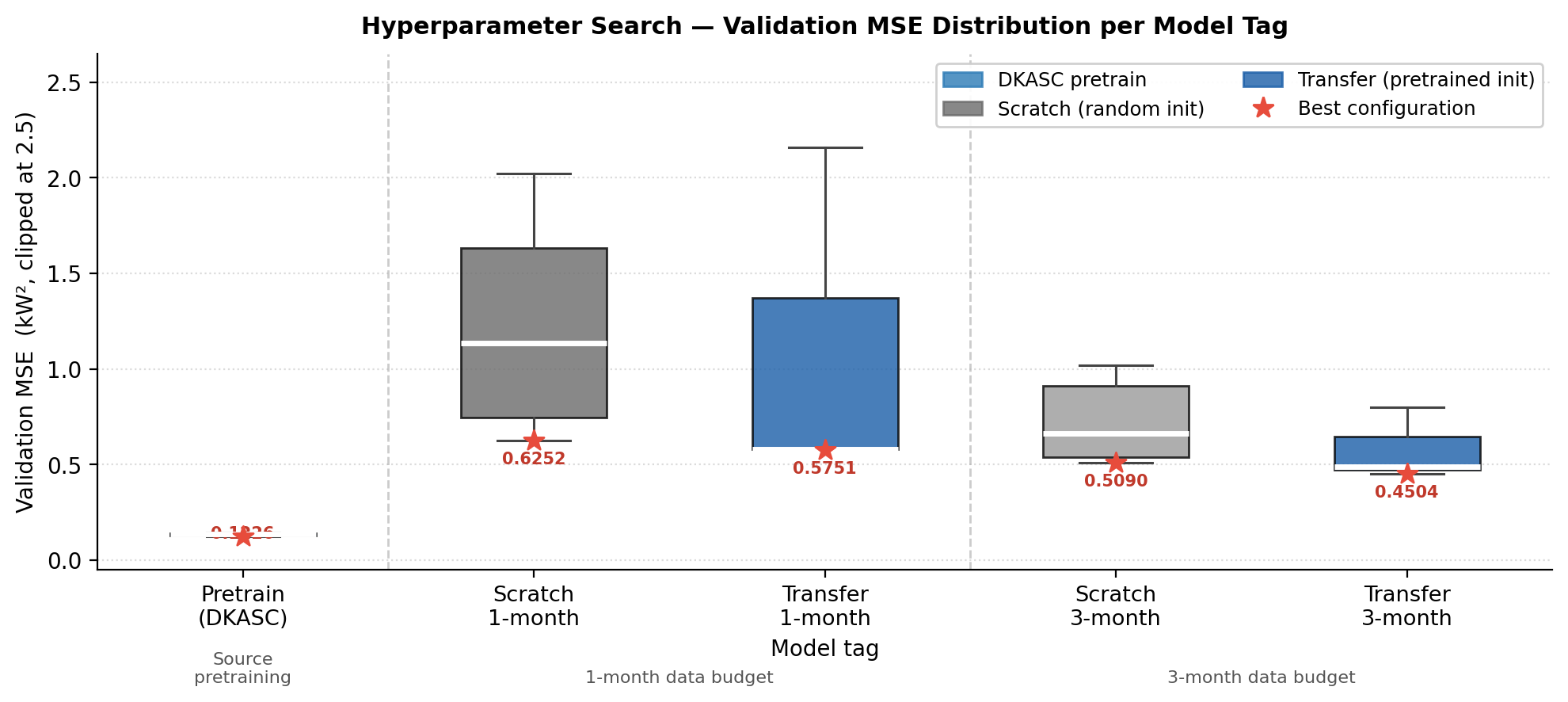}
    \caption{Distribution of validation MSE across hyperparameter configurations 
    per model tag (values clipped at 1.5~kW$^2$). Red stars mark 
    the best-performing configuration selected for each tag.}
    \label{fig:hparam_dist}
\end{figure}

\textbf{Scratch training.} For each data budget (one-month, three-month), an independent 
grid search is performed with random weight initialisation following the approach of 
Goodfellow et al.\ \cite{goodfellow2016deep}. Dropout \cite{srivastava2014dropout} 
and early stopping (patience = 10 epochs) prevent overfitting on the small target 
datasets.

\textbf{Transfer fine-tuning.} The pretrained backbone is loaded and all layers are 
unfrozen for fine-tuning on the target data, with a reduced learning rate 
($10^{-4}$ for one-month, $5\times10^{-4}$ for three-month) to prevent catastrophic 
forgetting \cite{yosinski2014transferable}. The best Transfer configurations 
(hidden size 32, 2 layers, dropout 0.2 in both cases) closely match the pretrained 
backbone, consistent with the hypothesis that source-learned representations are 
already well-suited to the target task.

\subsection{Conformal Quantile Regression}

After training, we apply CQR \cite{romano2019conformalized} to each of the four 
point-forecast--quantile model pairs (Scratch 1-month, Transfer 1-month, Scratch 
3-month, Transfer 3-month). Given quantile predictions 
$\hat{q}_\alpha^{lo}(x)$ and $\hat{q}_\alpha^{hi}(x)$ from the quantile head, 
the conformal nonconformity score for a calibration point $(x_i, y_i)$ is:

\begin{equation}
    s_i = \max\!\Bigl(\hat{q}_\alpha^{lo}(x_i) - y_i,\; y_i - \hat{q}_\alpha^{hi}(x_i)\Bigr).
    \label{eq:cqr_score}
\end{equation}

The conformal quantile $\hat{q}$ is the 
$\lceil (1-\alpha)(1+1/n_{cal})\rceil$-th smallest score over the $n_{cal}$ calibration 
points. At test time, the prediction interval for a new input $x$ is 

\begin{equation}
    \mathcal{C}(x) = \bigl[\hat{q}^{lo}(x) - \hat{q},\; \hat{q}^{hi}(x) + \hat{q}\bigr].
    \label{eq:cqr_interval}
\end{equation}

A 90\% nominal coverage level ($\alpha = 0.10$) is targeted. Because the calibration set is 
drawn from masked training data while the test set is clean, the exchangeability 
assumption underlying CQR is violated to a degree that depends on how much the 
mask distorts the calibration-set marginal distribution. This violation is precisely 
what we study.

\subsection{Evaluation Metrics}

Point-forecast quality is assessed by root mean squared error (RMSE), mean absolute 
error (MAE), and the forecast skill score (FSS) relative to a 24~h persistence baseline:

\begin{equation}
    \mathrm{FSS} = 1 - \frac{\mathrm{RMSE}_{\text{model}}}{\mathrm{RMSE}_{\text{persistence}}}.
    \label{eq:fss}
\end{equation}

Probabilistic quality is assessed by empirical coverage, mean interval width, and 
the mean pinball loss at the target quantile. An hourly coverage diagnostic (coverage 
stratified by hour of day) is used to localise the effects of structured missingness.

\section{Experiments and Results}
\label{sec:experiments}

\subsection{Point-Forecast Performance}

Table~\ref{tab:point_results} summarises point-forecast metrics for all four model 
variants on the 744-hour clean December 2023 test set.


\begin{table}[htbp]
\caption{Point-Forecast Evaluation on the December 2023 Test Set}
\begin{center}
\begin{tabular}{|l|c|c|c|c|}
\hline
\textbf{Model} & \textbf{Budget} & \textbf{RMSE (kW)} & \textbf{MAE (kW)} & \textbf{Skill} \\
\hline
Scratch  & 1-month & 0.659 & 0.375 & $-$0.033 \\
\hline
Transfer & 1-month & 0.503 & 0.274 & $+$0.212 \\
\hline
Scratch  & 3-month & 0.488 & 0.285 & $+$0.235 \\
\hline
Transfer & 3-month & 0.421 & 0.217 & $+$0.341 \\
\hline
\end{tabular}
\label{tab:point_results}
\end{center}
\end{table}

\begin{figure}[h]
    \centering
    \includegraphics[width=0.85\columnwidth]{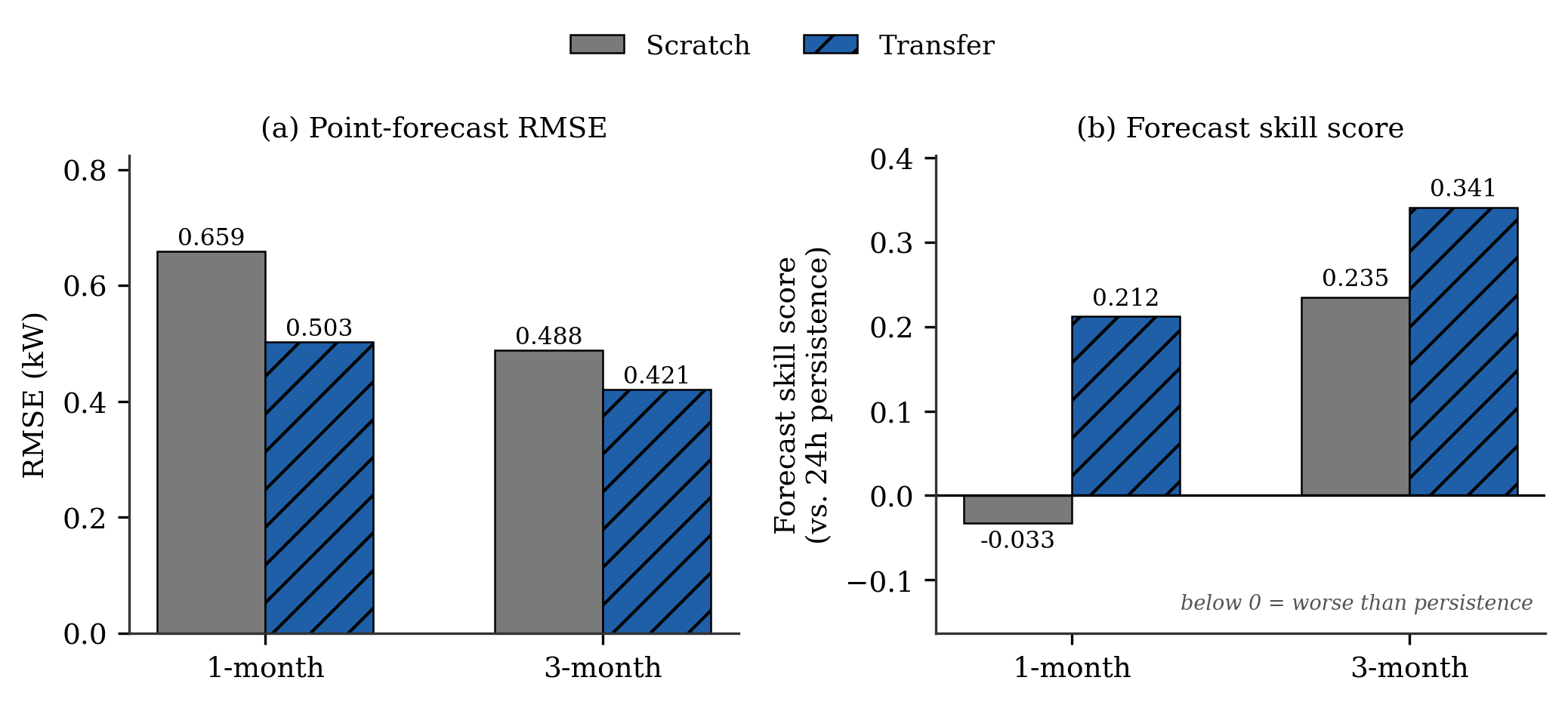}
    \caption{Point-forecast results. (a) RMSE for all four model variants; 
    (b) Forecast skill score relative to 24~h persistence. Transfer learning 
    flips the one-month model from skill-negative to skill-positive.}
    \label{fig:point_results}
\end{figure}

Two features of Table~\ref{tab:point_results} stand out. First, the transfer advantage 
is larger under tighter data budgets: Transfer reduces RMSE by 23.7\% at one month 
versus 13.7\% at three months, as expected since the pretrained prior matters most when 
the target set is too small to learn a good representation from scratch. Second, the 
one-month Scratch model achieves a skill score of $-$0.033-slightly worse than simply 
repeating yesterday's output, a benchmark requiring no learning at all-while Transfer 
reaches $+$0.212. This reversal reflects how damaging small data plus structured 
missingness is for scratch-trained networks: after masking, only a fraction of the 720 
training hours survive, biased toward nighttime and low-irradiance conditions, leaving 
the model poorly calibrated for bright midday. Fig.~\ref{fig:point_results} illustrates these findings graphically.

\subsection{Probabilistic Performance: CQR Calibration}

Table~\ref{tab:cqr_results} presents the CQR calibration results at the 90\% 
nominal coverage level.


\begin{table}[htbp]
\caption{CQR Probabilistic Evaluation (90\% Target Coverage)}
\begin{center}
\begin{tabular}{|l|c|c|c|c|}
\hline
\textbf{Model} & \textbf{Budget} & \textbf{Cov. (\%)$^{\mathrm{a}}$} & \textbf{Width (kW)} & \textbf{Pinball} \\
\hline
Scratch+CQR  & 1-month & 98.47 & 6.84 & 0.174 \\
\hline
Transfer+CQR & 1-month & 98.75 & 8.46 & 0.213 \\
\hline
Scratch+CQR  & 3-month & \textbf{84.17} & 2.31 & 0.064 \\
\hline
Transfer+CQR & 3-month & 94.31 & 1.98 & 0.054 \\
\hline
\multicolumn{5}{l}{$^{\mathrm{a}}$Bold indicates violation of the 90\% nominal coverage guarantee.}
\end{tabular}
\label{tab:cqr_results}
\end{center}
\end{table}

The one-month results reveal a pattern of severe overcoverage: both models produce 
intervals far wider than necessary, achieving empirical coverage of $\sim$98.5--98.8\% 
against a 90\% target. The Transfer+CQR model's conformal adjustment $\hat{q} = 3.51$~kW 
is notably large in absolute terms-on a 10~kWp nominal system, this represents 
a 35\% capacity expansion of each interval boundary. The root cause is that with 
only $n_{cal} \approx 144$ calibration points drawn from masked data, the nonconformity 
scores are highly variable; conformal theory dictates that $\hat{q}$ must then be 
very conservative to guarantee marginal coverage. The practical consequence is that 
one-month-trained CQR intervals, whether scratch or transfer, are too wide to be 
useful for day-ahead dispatch sizing.

The three-month results are more differentiated and tell a clear story. The 
Scratch+CQR model achieves only 84.17\% empirical coverage on the clean test 
set-a 5.8 percentage-point shortfall below the 90\% guarantee. This violation 
arises because the scratch model's calibration residuals are systematically biased: 
afternoon hours, which have the highest outage probability in the training mask, 
are underrepresented in the calibration set, so $\hat{q}$ is computed on a 
non-representative sample. When the test set presents clean afternoon observations, 
the scratch model's quantile predictions and the resulting $\hat{q}$ are both 
mis-calibrated.

Transfer+CQR, by contrast, achieves 94.31\% coverage with a mean interval 
width of 1.98~kW, 14\% narrower than the scratch counterpart (2.31~kW). This 
double benefit-better coverage \textit{and} tighter intervals-suggests that 
the transfer model's pretrained representations produce more consistent residuals 
across the solar day, reducing the sensitivity of $\hat{q}$ to the calibration 
set's missing-data bias. In other words, source-domain pretraining not only 
improves point predictions; it indirectly stabilises the conformal calibration 
process by narrowing the distribution of nonconformity scores.

\subsection{Hourly Coverage Diagnostic}

\begin{figure}[h]
    \centering
    \includegraphics[width=0.85\columnwidth]{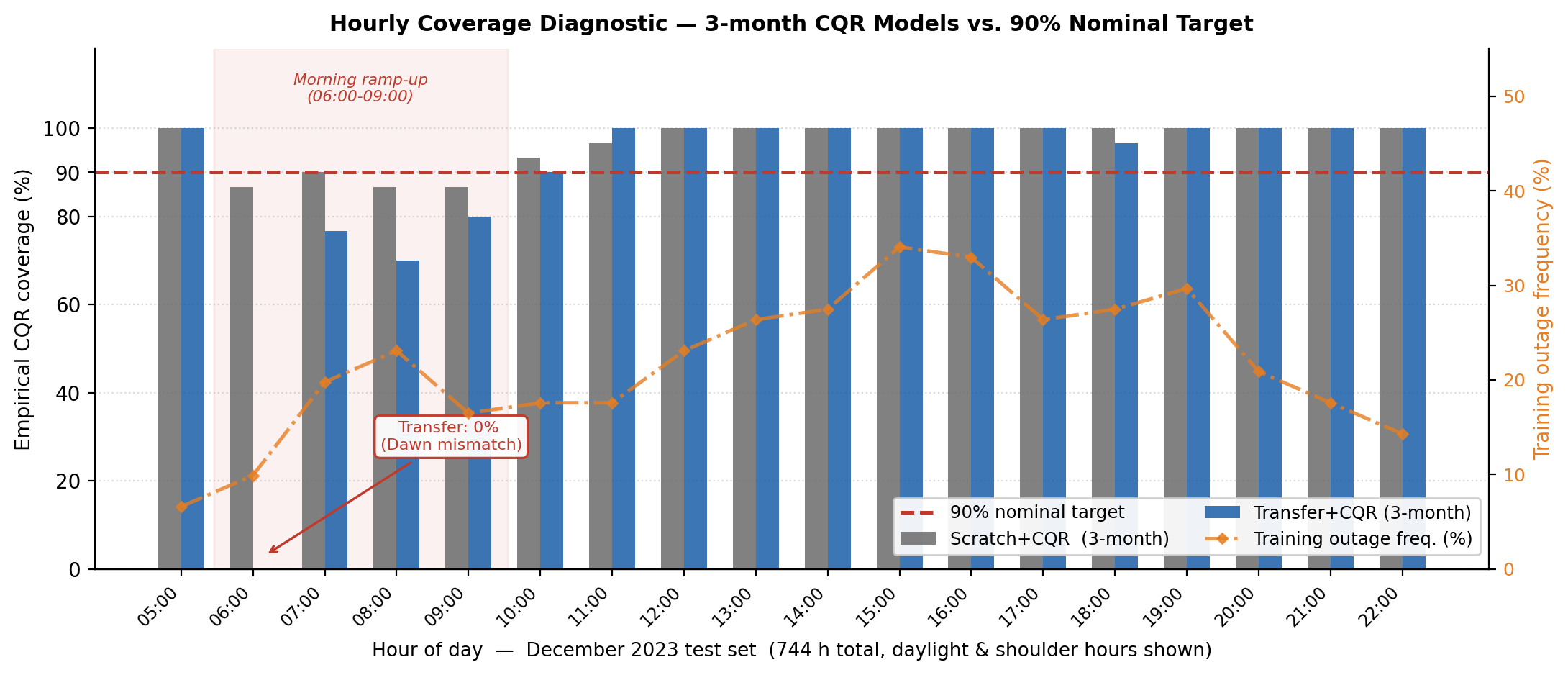}
    \caption{Hourly empirical coverage of Scratch+CQR and Transfer+CQR 
    (3-month models) on the December 2023 test set. Orange dash-dot line 
    indicates the training outage frequency at each hour. Red shading marks 
    morning ramp-up hours (06:00--09:00), the most problematic coverage window.}
    \label{fig:hourly_diag}
\end{figure}

To understand \textit{where} in the solar day the coverage shortfall of the 
three-month Scratch+CQR model originates, we computed per-hour empirical coverage 
on the December 2023 test set. Fig.~\ref{fig:hourly_diag} shows the per-hour 
empirical coverage for both models alongside the training outage frequency, 
and Table~\ref{tab:hourly_diag} summarises the most diagnostic hours.

\begin{table}[htbp]
\caption{Hourly Coverage Diagnostic (3-month models, Dec 2023)}
\begin{center}
\begin{tabular}{|c|c|c|c|}
\hline
\textbf{Hour} & \multicolumn{3}{|c|}{\textbf{Model Performance}} \\
\cline{2-4}
\textbf{} & \textbf{\textit{Train Outage Prob.}} & \textbf{\textit{Scratch Cov. (\%)}} & \textbf{\textit{Transfer Cov. (\%)}} \\
\hline
06:00 & 0.099 & 86.7 & 0.0$^{\mathrm{*}}$ \\
\hline
07:00 & 0.198 & 90.0 & 76.7 \\
\hline
08:00 & 0.231 & 86.7 & 70.0 \\
\hline
09:00 & 0.165 & 86.7 & 80.0 \\
\hline
13:00 & 0.341 & 100.0 & 100.0 \\
\hline
14:00 & 0.330 & 100.0 & 100.0 \\
\hline
\multicolumn{4}{l}{$^{\mathrm{*}}$No transfer prediction was made for this hour, resulting in 0\% coverage.}
\end{tabular}
\label{tab:hourly_diag}
\end{center}
\end{table}

The pattern reveals an important nuance. Hours 13:00--16:00, which have the 
highest outage frequency in the training mask (up to 34\%), paradoxically show 
100\% empirical coverage for both models. The reason is that these are also 
the peak irradiance hours, where both the point model and the quantile bounds 
are well constrained by the physical irradiance cycle; the conformal correction 
absorbs the worst residuals. In contrast, morning hours 07:00--09:00, which 
have moderate outage frequencies (17--23\%) but represent the irradiance 
ramp-up phase, show below-target coverage for both models, with the Transfer 
model performing worse at those specific hours.

A striking anomaly appears at 06:00, where the Transfer model achieves zero 
empirical coverage in December. This hour corresponds to pre-sunrise in 
Bangladesh during the December solstice, when irradiance is near-zero but 
the transition from darkness to dawn is sharper than in Alice Springs at the 
same time of year. The DKASC pretraining has apparently embedded a slightly 
different dawn-timing prior that manifests as miscalibration at this marginal 
hour. This suggests that climate-regime mismatch at dawn and dusk transitions 
is a key failure mode to watch for when transferring between Southern and 
Northern hemisphere sites.

Overall, the hourly diagnostic confirms that the global coverage shortfall 
in Scratch+CQR (84.17\%) originates predominantly from morning irradiance-ramp 
hours, where training-data gaps under the outage mask are most damaging to 
calibration-set representativeness.

\section{Discussion}
\label{sec:discussion}

Transfer helps more at one month (23.7\% RMSE reduction) than at three months
(13.7\%) because pretrained priors matter most when target data are scarcest:
after masking, fewer than 50 midday training hours remain in the one-month
window-too few for an LSTM to learn the irradiance-to-power mapping from
scratch, as reflected in a skill score of $-$0.033 that fails to beat
persistence.

This scarcity also undermines the exchangeability assumption behind our
conformal guarantees \cite{vovk2005algorithmic,angelopoulos2022gentle}:
load-shedding strips high-irradiance afternoon hours from the calibration
set but not the test set, producing the shift described in
\cite{tibshirani2019conformal} and observed as 84.17\% coverage against a
90\% target for Scratch+CQR. Transfer mitigates this indirectly, since
pretrained features give more consistent nonconformity scores across the
solar day, yielding a better-estimated $\hat{q}$ from an incomplete
calibration set.

Practically, this makes the system useful from month one (skill score
+0.212 for Transfer), and by month three the 94.31\% coverage interval lets
an EMS treat its lower bound as a conservative irradiance forecast for
triggering diesel backup-though the dawn-hour miscalibration in
Table~\ref{tab:hourly_diag} suggests a clear-sky correction is needed for
early-morning ramp-up decisions.

\begin{figure}[h]
    \centering
    \includegraphics[width=0.65\columnwidth]{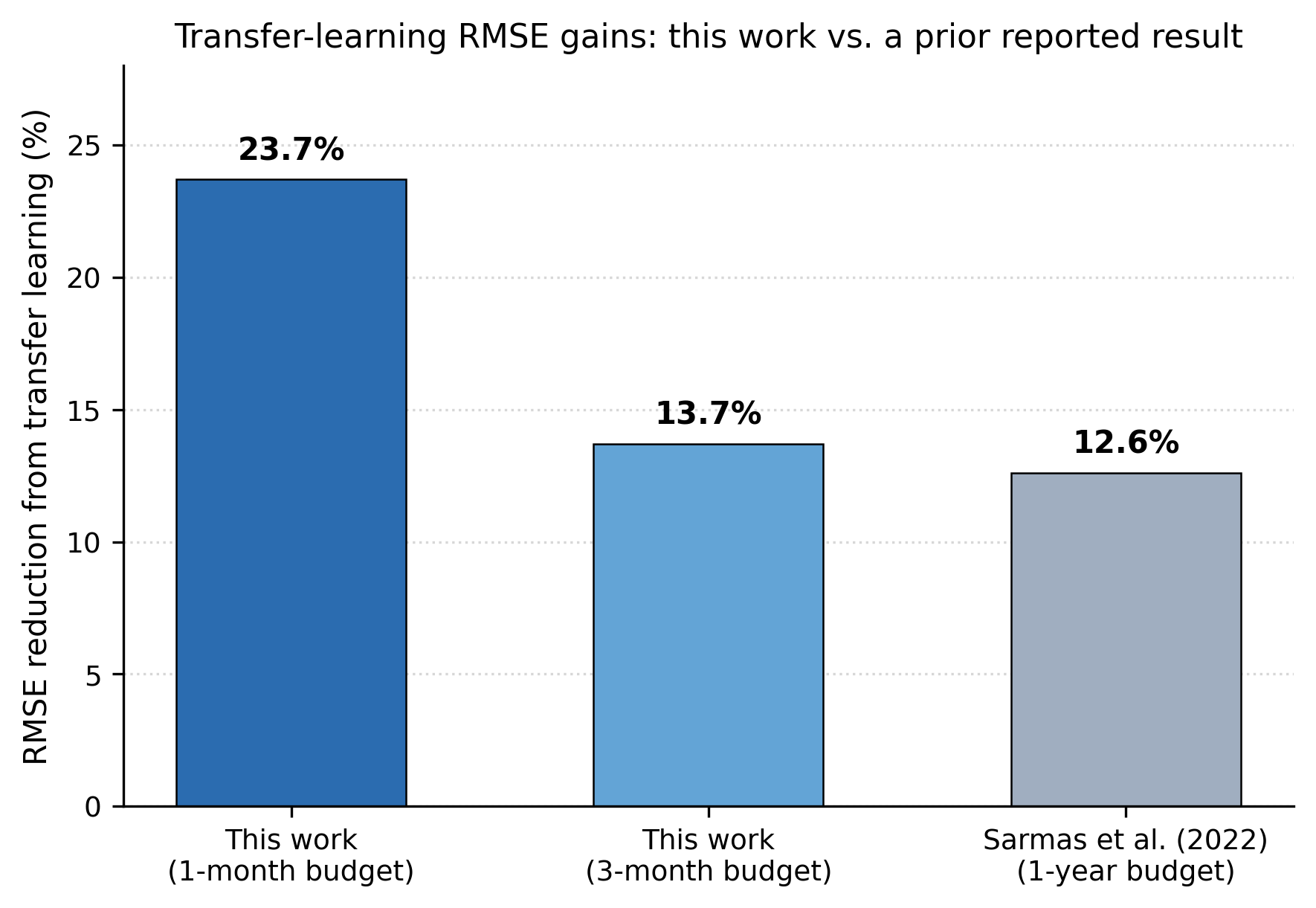}
    \caption{RMSE reduction from transfer learning: this work (23.7\% at 
    one month; 13.7\% at three months) vs.\ 12.6\% reported by Sarmas et al.\ 
    \cite{sarmas2022transfer} with one year of data (indicative, not a 
    controlled comparison; sites/scales/budgets differ).}
    \label{fig:tl_comparison}
\end{figure}

Fig.~\ref{fig:tl_comparison} places our gains alongside the only prior 
PV transfer-learning result we could verify in comparable, scale-free 
form: Sarmas et al.\ \cite{sarmas2022transfer} report 12.6\% RMSE 
improvement from transfer with a full year of data. Our one-month model 
nearly doubles that (23.7\%) with far less, harder data, supporting our 
claim that transfer's relative value grows as the budget shrinks. Other 
TL-for-PV papers report only absolute RMSE/MAE at incompatible scales, 
so we did not force them into the same chart.

Key limitations: the Bangladesh data are simulated (no real soiling,
shading, or sensor noise), the missingness model uses aggregated BPDB
statistics rather than site-specific logs, and a single Australian source
site may transfer less effectively than a more diverse or regionally closer
portfolio. Replacing the fixed calibration window with online adaptive
conformal methods \cite{moradi2025enhanced,renkema2024conformal} is a
natural next step. We regard reliance on simulated target-domain data as 
the most consequential limitation, leaving these results closer to a 
proof-of-concept than a field-validated system. No public, measured 
Bangladesh rooftop PV dataset appears to exist; the closest available 
alternative is a real, measured 10~kWp rooftop dataset from IIEST Shibpur, 
West Bengal, India ($\sim$1.2\textdegree{} from our Dhaka target, same 
monsoon regime) \cite{chakraborty2023computational}. We have not re-run 
our pipeline on it-this needs a site-specific outage mask and re-fitted 
backbone-but flag it as the concrete next step for external validation.

\section{Conclusion}
\label{sec:conclusion}

We studied solar PV forecasting for newly deployed microgrids where load-shedding-driven
measurement gaps limit both data volume and continuity. Three findings stand out. First,
transfer learning from a data-rich Australian source yields meaningful point-forecast
gains at one- and three-month budgets, and is critical at one month, where scratch
training fails to beat persistence. Second, structured missingness propagates into
conformal calibration: the three-month Scratch+CQR model undercovers by nearly six
points, while transfer initialisation dampens this, achieving near-nominal 94.3\%
coverage with 14\% narrower intervals. Third, hourly diagnostics identify dawn-hour
ramp-up as the most vulnerable part of the solar day, a concrete target for future work.

Crucially, these benefits are not merely additive: pretrained representations
stabilise the conformal calibration process in a way neither ingredient achieves
independently. For the growing number of microgrids deployed in data-scarce,
grid-unreliable regions across South and Southeast Asia and Sub-Saharan Africa,
this framework provides a practical, deployment-ready forecasting stack from day one.

\section*{Acknowledgment}

The authors gratefully acknowledge the support of CRITS, Green University of Bangladesh in facilitating this research.

\bibliographystyle{IEEEtran}
\bibliography{references}
\end{document}